\documentclass{llncs}
\usepackage{graphicx}
\usepackage{amsmath}
\usepackage{amssymb}
\usepackage{amstext}
\usepackage{ifthen}
\usepackage{float}

\def\dst{\displaystyle}

\def\ADD#1#2{#1+#2}

\def\MUL#1#2{#1*#2}
\def\DIV#1#2{{#1 \over #2}}

\def\esp{\hspace*{.5in}}
\begin{document}
\title{Avoiding the Bloat with
Stochastic Grammar-based Genetic Programming}
\author{Alain Ratle\inst{1} and Mich\`ele Sebag\inst{2}}
\institute{
LRMA- Institut Sup\'erieur de l'Automobile et des Transports 
58027 Nevers France\\
Alain.Ratle\_isat@u-bourgogne.fr
\and
LMS CNRS UMR 76-49, Ecole Polytechnique 
91128 Palaiseau France \\
Michele.Sebag@Polytechnique.fr
} 
\maketitle

\begin{abstract}
The application of Genetic Programming to the discovery of empirical laws is often 
impaired by the huge size of the search space, and consequently by the computer resources needed. 
In many cases, the extreme demand for memory and CPU is due to the massive growth of 
non-coding segments, the introns. The paper presents a new program evolution framework which 
combines distribution-based evolution in the PBIL spirit, with grammar-based genetic 
programming; the information is stored as a probability distribution on the grammar rules, 
rather than in a population. Experiments on a real-world like problem show that this 
approach gives a practical solution to the problem of intron growth.
\end{abstract}

\section{Introduction}
This paper is concerned with the use of Genetic Programming 
(GP) \cite{Koza,Banzhaf:book} for the automatic discovery of empirical 
laws. Although GP is widely used for symbolic regression 
\cite{McKay95,Duffy99}, it suffers from two main limitations. 
One first limitation is that canonical GP offers no way to 
incorporate domain knowledge besides the set of operators, despite 
the fact that the knowledge-based issues of Evolutionary 
Computation are widely acknowledged \cite{Radcliffe,Janikow}.

In a previous work \cite{Ratle-Sebag:PPSN2000} was described a hybrid
scheme combining GP and context free grammars (CFGs). First 
investigated by Gruau \cite{GruauPhD} and Whigham \cite{Whigham95}, 
CFG-based GP allows for expressing and enforcing syntactic 
constraints on the GP solutions. We applied CFG-based GP to enforce
the dimensional consistency of empirical laws. Indeed, in virtually
all physical applications, the domain variables are labelled 
with their dimensions (units of measurement), and the solution 
law must be consistent with respect to them (seconds and meters
should not be added). Dimensional consistency allows for 
massive contractions of the GP search space; it significantly
increases the accuracy and intelligibility of the empirical 
laws found.

A second limitation of GP is that it requires huge amounts of 
computational resources, even when the search space is properly
constrained. This is blamed on the bloat phenomenon, resulting from 
the growth of non-coding branches ({\em introns}) in the GP individuals
\cite{Koza,LangdonPoli97a}. The bloat phenomenon adversely affects GP in two
ways; on one hand, it might causes the early termination of the GP runs 
due to the exhaustion of available memory; on the other hand, it significantly 
increases the  fitness computation cost.

In this paper a new GP scheme addressing the bloat phenomenon is presented,
which combines CFG-based GP and distribution-based evolution.
In distribution-based evolution, an example of which is 
PBIL \cite{Baluja:ICML95}, the genetic pool is coded as a distribution on the
search space; in each generation, the population is generated from the 
current distribution; and the distribution is updated from the
best (and possibly the worst) individuals in the current population.

In this new scheme, termed SG-GP (for {\em 
Stochastic Grammar-based GP}), 
the distribution on the GP search space is represented as  a stochastic
grammar. It is shown experimentally that this scheme prevents the 
uncontrolled growth of introns. This result offers new hints into the 
bloat phenomenon.

The paper is organized as follows. The next section briefly summarizes context-free
grammars (CFGs) and CFG-based GP, in order for the paper to be self contained.
The principles of Distribution-based evolution are presented in section \ref{DE},
and related works are discussed \cite{SalusSchmid98}.
Stochastic Grammar based GP is detailed in Section \ref{SG}. 
An experimental validation of SG-GP on real-world problems 
is reported in Section \ref{Expe},
and the paper ends with some perspectives for further research.

\section{CFG-based GP}\label{SectBNF}
\subsection{Context Free Grammars}
A context free grammar describes  the admissible constructs of a language by 
a 4-tuple $\{S,N,T,P\}$, where $S$ is the start symbol, $N$ the set 
of non-terminal symbols, $T$ the set of terminal symbols, and 
$P$ the production rules.
Any expression is iteratively built up from the start symbol 
by rewriting non-terminal symbols into one of 
their derivations, as given by the production 
rules, until the expression contains terminals only.
Fig. 1 shows the CFG 
 describing the polynoms of variable $X$, to be compared with the
standard GP description from the  node set ${\mathcal N} = \{ +, \times \}$ and terminal set ${\mathcal T} = \{ X, {\mathcal R} \}$~:
{\small\begin{center}
$\begin{array}{r l}
N &=~\{<exp>,~<op>,~<var>\} \\
T &=~\{+,~\times,~X,~\mathcal R~\} \esp  \mbox{//$\mathcal R$ stands for any real-valued constant}\\
P &=~\left \{ \begin{array}{cl}
S     &:=~ <exp>~; \\
<exp>  &:=~~<op>~<exp>~ <exp>~~|~<var> ~;\\
<op> &:=~+~|~\times~;\\
<var>  &:=~X~|~\mathcal R~; \\
\end{array} \right\} \\
\end{array}
$\smallskip\\
{Fig.1. Context Free Grammar for polynoms of any degree of variable $X$}
 \end{center}
}
Note that {\em non-terminals} and {\em terminals} have different 
meanings in GP and in CFGs. GP terminals (resp. non-terminals) stand
for domain variables and constants (resp. operators). CFGs terminals
comprise domain variables, constants, and operators.
\subsection{CFG-based GP}
On one hand, CFGs allow one to express problem-specific constraints
on the GP search space. On the other hand, the recursive application 
of derivation rules allows the build up of a 
{\em derivation tree} (Fig. 2), which can be thought of 
as an alternative representation for the expression tree. In this case,
$S$ stands for the start symbol, $E$ for an expression, $v$ for a variable 
and $Op$ for an operator.

\begin{center}
\includegraphics*[width=0.6\textwidth]{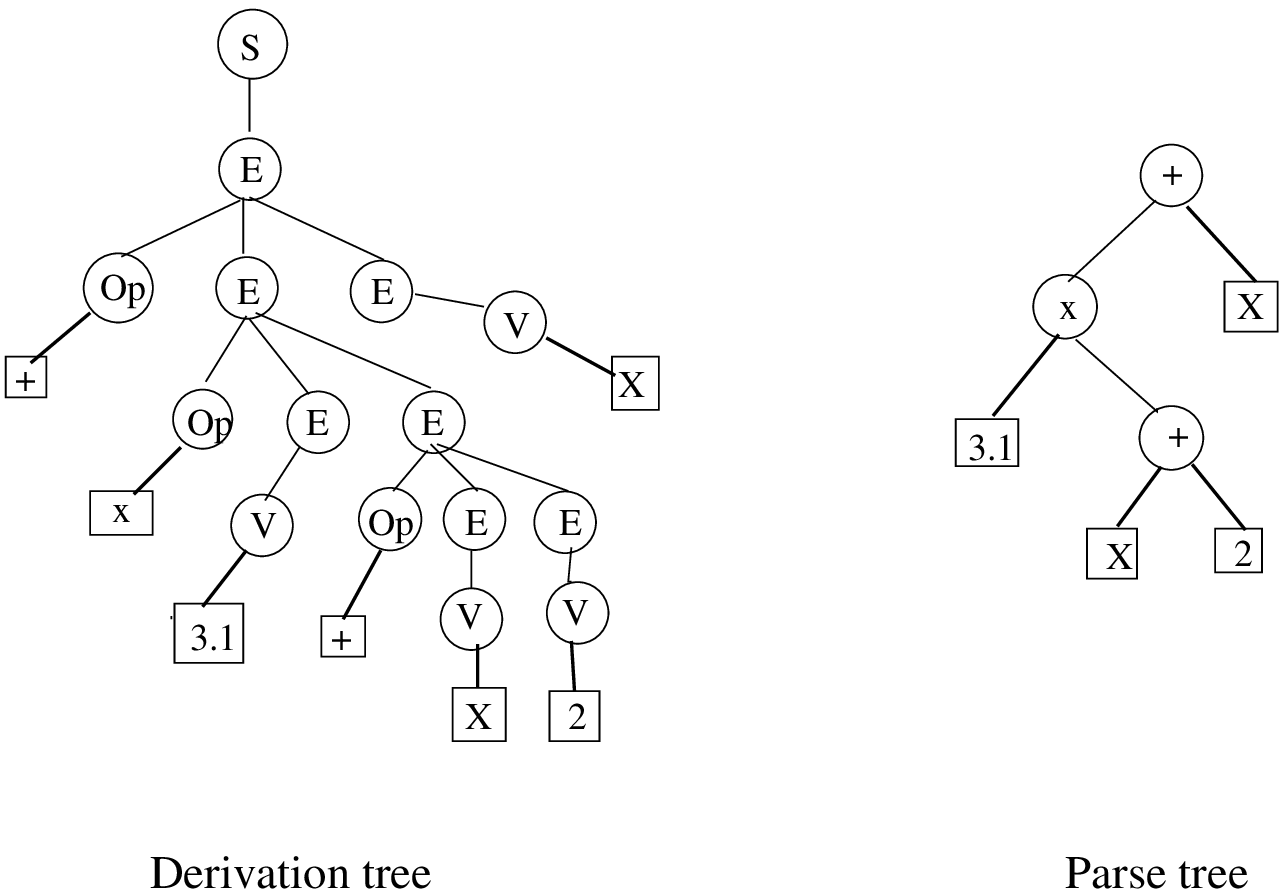}\\
{Fig.2. Derivation tree and Corresponding Parse tree}
\end{center}

Derivation trees can be manipulated using evolution operators.
In order to ensure that CFG-compliant offspring are produced 
from  CFG-compliant 
parents, crossover is  restricted to swapping subtrees 
built on the same non-terminal symbol; mutation replaces a subtree
by a new tree built on the same symbol
\cite{GruauPhD,Whigham95}. These 
restrictions are quite similar to that of Strongly Typed GP \cite{Montana95}.

\subsection{Dimensionally-aware GP}\label{CFG-GP}
As mentioned in the introduction, the discovery of empirical laws 
is easier when units of measurement are taken into account.
These units can be expressed with resprect to a set of elementary units, and 
represented as vectors
 (e.g. {\em Newton = mass $\times$ length $\times$ time $^{-2}$} 
is represented as vector $[1,1,-2]$). 
Restricting ourselves to a finite number of compound units, 
a non-terminal symbol is associated to any allowed compound unit. The associated
derivation rule describes all possible ways for generating an expression 
of the given unit. An automatic grammar
generator takes as input the elementary units and the set of compound
units allowed, and produces the CFG describing all dimensionally 
consistent expressions in the search space\footnote{
The production rule associated to the start symbol specifies the 
unit of the sought solution; it can also enforce the shape of the 
solution, according to the expert guess.}. Although the CFG size is
exponential, enforcing these restrictions linearly increases the crossover
complexity in the worst case, and does not modify the mutation complexity.

Compared to most CFGs used  in the GP literature \cite{GruauPhD}, 
the dimensional-CFG is huge (several hundreds
non-terminal symbols, several thousands of derivations). The inefficiency
of CFG-GP in this frame, already reported by \cite{Ryan98}, was blamed
on the initialization operator. This drawback was addressed by a specific,
constrained grammar-based initialization process, building a CFG-compliant
{\em and sufficiently diverse} initial population.
 The core of the procedure 
is a two-step process: a) for any given non-terminal symbol,
all derivations compatible with the  maximum tree-depth prescribed
(ensuring that the final expression will have admissible size) are determined;
 b) the non-terminal symbol at hand is rewritten by uniformly selecting
one compatible derivation (see \cite{Ratle-Sebag:PPSN2000} for more details).

\section{Distribution-based Evolution}\label{DE}
Contrasting with genetic evolution, distribution-based evolution 
deals with a high-level (intentional) 
description of the best individuals encountered so far, 
as opposed to the (extensional) description given by the current population
itself. This intentional description is a probability distribution on the 
solution space, which is updated according to a set of rules.

As far as we know, the first algorithm 
resorting to distribution-based evolution 
is {\em Population-based Incremental Learning (PBIL)}
\cite{Baluja:ICML95}, concerned with optimization in $\{0,1\}^n$. 
In this scheme, distribution $\cal M$ is represented as an 
element of $[0,1]^n$, initialized to ${\cal M}_0 = (.5, \ldots, .5)$.\\
At generation $t$, ${\cal M}_t$ is used to generate the population
from scratch, where the probability for any individual $X$ to have its $i$-th
bit set to 1 is given as the $i$-th component of ${\cal M}_t$.
The best individual $X_{best}$  in the current population is used 
to update ${\cal M}_t$ by relaxation\footnote{Other 
variants use the best two individuals, and possibly the worst one too,
to update the distribution.}, with 
\[ {\cal M}_{t+1} = (1 - \epsilon) {\cal M}_t + \epsilon X_{best}\]
${\cal M}_t$  is also randomly perturbed (mutated) to avoid premature convergence.

This scheme has been extended to accommodate different
distribution models and non-binary 
search spaces (see \cite{Sebag:PPSN98,Larranaga} among others).

Distribution-based evolution has been extended to GP through the 
{\em Probabilistic Incremental Program Evolution (PIPE)}
system \cite{SalusSchmid98}. The distribution on the GP search space is 
represented as a Probabilistic Prototype Tree (PPT); in each PPT node 
stand the probabilities for selecting any variable and operator
in this node. After the current individuals have been constructed and
evaluated, the PPT is biased toward the current best and the best-so-far
individuals. 
One feature of the PIPE system is that the PPT grows deeper and wider
along evolution, depending on the size of the best trees, since the 
probabilities of each variable/operator have to be defined for 
each possible position in the tree.

\section{Stochastic Grammars-based GP (SG-GP)}\label{SG}

\subsection{Overview}
Distribution-based evolution actually involves three components:
the representation (model) for the distribution; the exploitation of 
the distribution in order to generate the current population, 
which is analogous in spirit to the genetic initialization
operator; the update mechanism, 
evolving the distribution according to the most remarkable 
individuals in the current population.

In CFG-GP, initialization proceeds by iteratively rewriting each
non-terminal symbol; this is done by selecting a derivation in the 
production rule associated to the current non-terminal symbol 
(e.g. $<exp>$ is either rewritten as a more complex expression, 
$<op> <exp> <exp>$, or a leaf $<var>$, Fig. 1). 
The selection is uniform (among the  derivations compatible
with the maximum tree size allowed, see Section
\ref{CFG-GP}). It comes naturally to encode the experience gained
from the past generations, by setting selection probabilities on the 
derivations. 

{\bf Representation}. 
The distribution over the GP search space is represented
as a stochastic grammar: each derivation $d_i$ 
in a production rule is attached a weight $w_i$, and the chances for
selecting derivation $d_i$ are proportional to $w_i$. 

{\bf Exploitation}.
The construction of the individuals from the 
current stochastic grammar 
is inspired from the  CFG-GP initialization procedure.
For each occurrence of a non-terminal symbol, all 
admissible derivations are  
determined from the maximum tree size allowed and the position
of the current non-terminal symbol as in \cite{Ratle-Sebag:PPSN2000}; the 
selection of the derivation $d_i$ is done with 
probability $p_i$, where
\begin{equation}
p_i =
\begin{cases}
\dst w_i \over \dst \sum\limits_{k \in \text{admissible derivs.}}\dst  w_k & 
\text{ if $d_i$ is an admissible derivation}\\
0 &\text{otherwise}\\
\end{cases}
\end{equation}
This way, weights $w_i$ need not be normalized.

{\bf Distribution update}.
After all individuals in the current population have been evaluated, 
the probability distribution is updated from the $N_b$ best and $N_w$ worst
individuals according to the following rules: for each derivation $d_i$,
\begin{itemize}
\item Let $b$ denotes the number of individuals among the $N_b$ best
individuals that carry derivation $d_i$; weight $w_i$ is multiplied 
by $(1 + \epsilon)^b$;
\item Let $w$ denotes the number of individuals among the $N_w$ worst
individuals that carry derivation $d_i$; weight $w_i$ is divided 
by $(1 + \epsilon)^w$;
\item Last, weight $w_i$ is mutated with probability $p_m$; the mutation
either multiplies or divides $w_i$ by factor $(1+\epsilon_m)$. 
\end{itemize}
All $w_i$ are initialized to 1. Note that it does not make sense
to have them normalized; they must be locally renormalized before use, 
depending on the current set of admissible derivations.

This distribution-based GP, termed SG-GP, involves five
parameters besides the three standard GP parameters (Table 1).
\begin{table}[ht]
\centering
\caption{Parameters of Stochastic Grammar-based Genetic Programming}
\label{ParamGGILP}
\begin{tabular}{|l|l|}
\hline
Parameter & Definition \\
\hline
\multicolumn{2}{|c|}{Parameters specific to SG-GP}\\
\hline
$N_b$   & Number of best individuals for probability update \\
$N_w$   & Number of worst individuals for probability update \\
$\epsilon$ &  Learning rate \\
$p_m$      & Probability of mutation \\
$\epsilon_m$ & Amplitude of mutation \\
\hline
\multicolumn{2}{|c|}{Canonical GP parameters}\\
\hline
$P$   & Population size \\
$G$   & Maximum number of generations \\
$D_{max}$   & Maximum derivation depth \\
\hline
\end{tabular}
\end{table}

\subsection{Scalar and Vectorial SG-GP}
In the above scheme, the genetic pool is represented by a 
vector $\cal W$, coding all derivation weights for all production rules.
The storage of a variable length population is replaced by the 
storage of a single fixed size vector; this is
in sharp contrast with canonical GP, and more generally, with 
all evolutionary schemes dealing with variable size individuals. 

One limitation of this representation is that it induces a total order
on the derivations in a given production rule. However, it might happen that
derivation $d_i$ is more appropriate than $d_j$ in higher levels of 
the GP trees, whereas $d_j$ is more appropriate in the bottom of the trees.

To take into account this effect, a distribution vector 
${\cal W}_i$ is attached to the $i$-th level of the GP trees
($i$ ranging from 1 to $D_{max}$). 
This scheme is referred to as {\em Vectorial SG-GP}, 
as opposed to the previous scheme of {\em Scalar 
SG-GP}.

The distribution update in Vectorial SG-GP is modified in a straightforward
manner; the update of distribution ${\cal W}_i$ is only based on the
derivations actually occurring at the $i$-th level among the best and 
worst individuals in the current population.

\subsection{Investigations on intron growth}

It have often been observed experimentally that the proportion of introns in the GP material  
grows exponentially along evolution \cite{NBF97}. As already
mentioned, the intron growth is undesirable as it drains out 
the memory resources, and increases the total fitness computation cost. 

However, it was also observed that pruning the introns in each
generation significantly decreases the overall GP performances
\cite{Koza}. Supposedly, introns protect good building blocks from the
destructive effects of crossover; as the useful part of the genome is
condensed into a small part of the individual, the probability
for a crossover to break down useful sections is reduced by the apparition
of introns.

Intron growth might also be explained from the structure of 
the search space \cite{LangdonPoli97a}. Consider all genotypes (GP trees) 
coding a given phenotype (program). There exists a lower bound on the 
genotype size (the size of the shortest tree 
coding the program); but there exists no
upper bound on the genotype size (a long genotype can always 
be made longer by the addition of introns). Since there are many more 
long genotypes than short ones, longer genotypes will be selected 
more often than shorter genotypes (everything else being equal, i.e. 
assuming that the genotypes are equally fit)\footnote{ To prevent or reduce this intron
growth, a parsimony pressure might be added to the fitness 
function \cite{parsimony}; but the relative importance of the actual 
fitness and that of the parsimony term must be adjusted carefully. And the 
optimal trade-off might not be the same for the beginning and the end of 
evolution.}.\\
Last, intron growth might also be a mechanical effect of evolution.
GP crossover facilitates the production of larger and larger 
trees: on one hand, the offspring average size is equal to the parent 
average size; on the other hand, short size offspring  usually are 
poorly fit; these remarks together explain why the individual size
increases along evolution.

Since the information transmission in SG-GP radically differs from that
in GP, as there exists no crossover in SG-GP, there should be no occasion
for building longer individuals, and no necessity for protecting
the individuals against destructive crossover.

Experiments with SG-GP are intended to assess the utility of introns. 
If the intron growth is beneficial {\em per se}, then 
either SG-GP will show able to produce introns | or the overall 
evolution results will be significantly degraded. If none of these
eventualities is observed, this will suggest that the role of introns 
has been overestimated.

\section{Experimental validation}\label{Expe}

\subsection{Test problem}
The application domain selected for this  study is related 
to the identification of rheological models. These problems 
have important applications in the development 
of new materials, especially for polymers and composite 
materials \cite{Ward85}. 
The target empirical law corresponds to the Kelvin-Voigt model, which 
consists of a spring and a dashpot in parallel (Fig. 3). When a constant
force is applied, the response (displacement-time relation) is
\[
u(t) = {F \over K} \left( 1-e^{\dst -Kt \over \dst C}  \right)
\]
Fitness cases (examples) are generated using random values of the 
material parameters $K$ and $C$ and loading $F$. 
The physical units for the domain
variables and for the solution are expressed with respect to the 
elementary $mass$, $time$ and $length$ units (Table 2).
Compound units are restricted as the exponent
for each elementary unit ranges in $\{-2,-1,0,1,2\}$. The dimensional
grammar is generated as described in Section~\ref{CFG-GP}, with 
125 non-terminal symbols and four operators (addition, 
multiplication, protected division and exponentiation). The
grammar size is about 515 k.

\begin{center}
\begin{tabular}{ccc}
\begin{tabular}{c}
\includegraphics*[width=.4\linewidth,height=1.in]{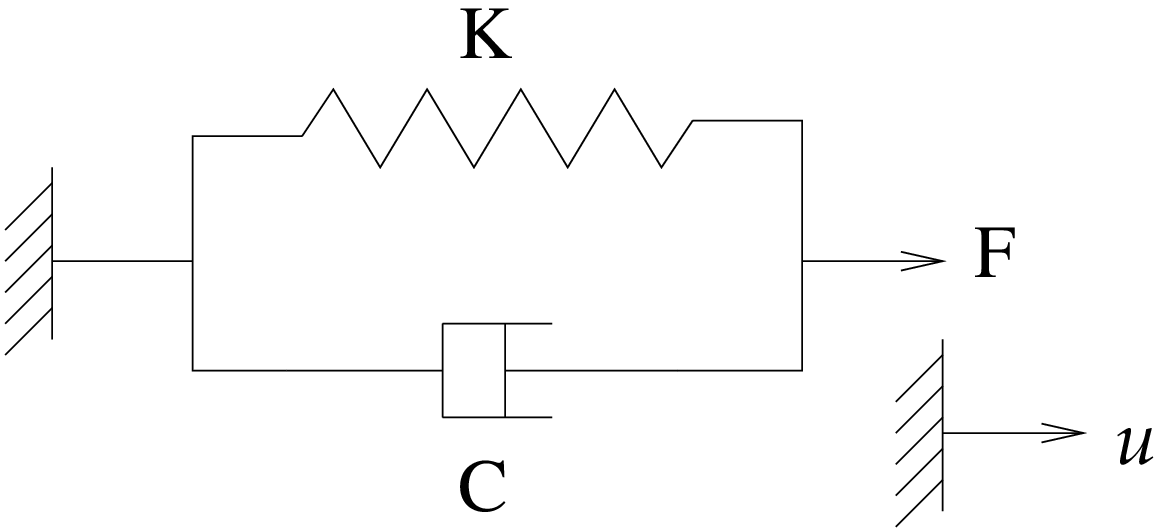}
\end{tabular} & \hspace{.2in} &
{\small
\begin{tabular}{|l|c|c|c|}
\hline
\multicolumn{4}{|c|}{Physical units}\\
\hline
Quantity & mass & length & time \\
\hline
\multicolumn{4}{|c|}{\it Variables} \\
\hline
$F$ (Force)                     &  +1 &  +1 & --2  \\
$K$ (Elastic elements)          &  +1 &   0 & --2  \\
$C$ (Viscous elements)          &  +1 &   0 & --1  \\
$t$ (time)                      &   0 &   0 &  +1  \\
\hline
\multicolumn{4}{|c|}{\it Solution} \\
\hline
$u$ (displacement)              &   0 &  +1 &   0 \\
\hline
\end{tabular}}\\
~\\
Fig. 3 Kelvin-Voigt Model & & Table 2.  Physical Units\\
\end{tabular}
\end{center}

\subsection{Experimental setting}
SG-GP is compared\footnote{Due to space 
limitations, the reader interested in 
the comparison of CFG-GP with GP is referee to \cite{Ratle-Sebag:PPSN2000}.}
 with standard elitist GP.
The efficiency of SG-GP is assessed with respect to the 
quality of solutions, and in terms of memory use. All results are 
averaged on 20 independent runs.

GP and SG-GP parameters are set according to a few preliminary experiments
(Table 3). Canonical GP is known to work better with large populations 
and small number of generations \cite{Koza}. 
Quite the contrary, SG-GP works better with a small
population size and many generations.
In both cases, evolution is stopped after 2,000,000 fitness evaluations.
\begin{center}
{\small\begin{tabular}{|c|c|c|c|}
\hline
\multicolumn{2}{|c|}{SG-GP} & \multicolumn{2}{|c|}{GP }\\
\hline
Parameter & Value & Parameter & Value\\
\hline
Population size                 & 500   & Population size               & 2000 \\
Max. number of generations      & 4000  & Max. number of generations    & 1000 \\
$N_b$   &  2                    & P(crossover) & 0.9    \\
$N_w$   &  2                    & P(mutation) &  0.5 \\
$\epsilon$ & 0.001              & Tournament size & 3 \\
$P_m$      & 0.001 & &\\
$\epsilon_m$ & 0.01 & & \\
\hline
\end{tabular}}\smallskip\\
{Table 3. Optimization parameters}
\end{center}

\subsection{Experimental results and parametric study}
\begin{center}
\begin{tabular}{ccc}
\includegraphics*[width=.4\linewidth]{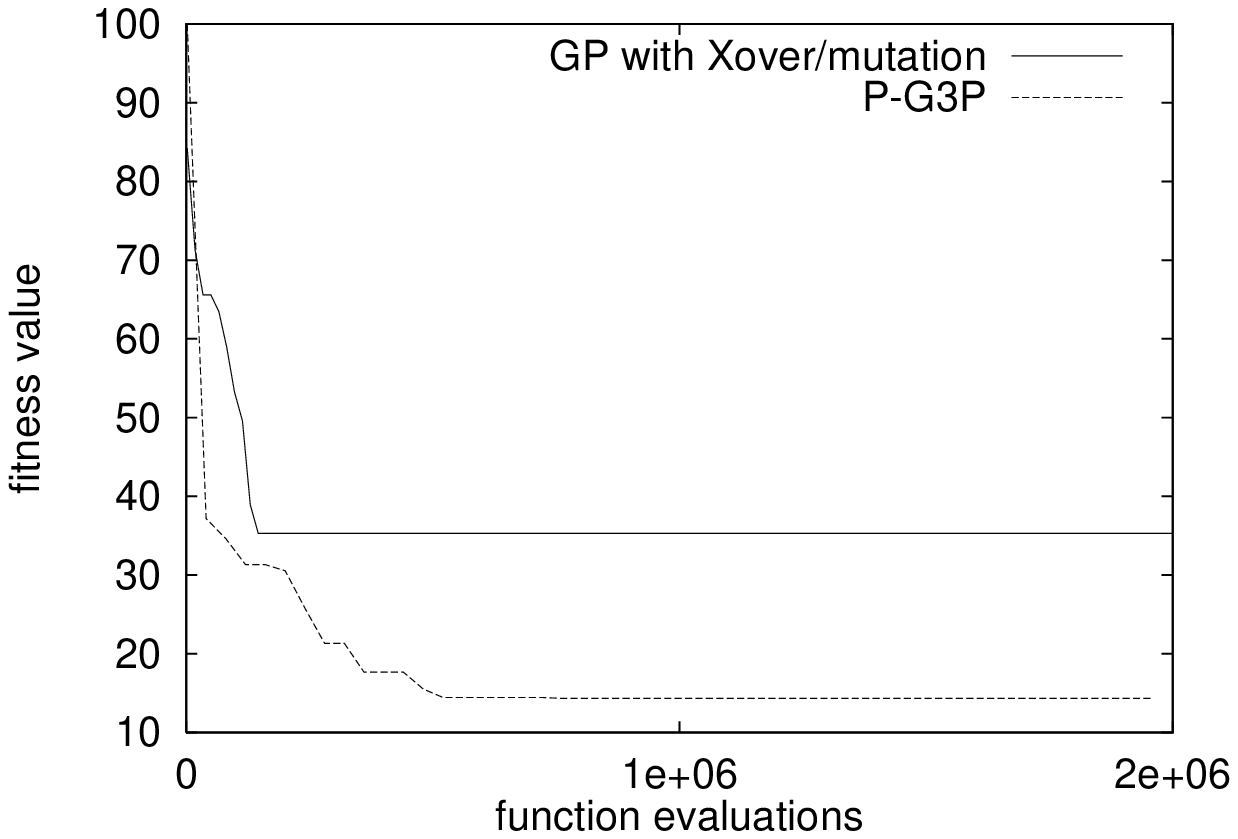} &\hspace{.2in} &
\includegraphics*[width=.4\linewidth]{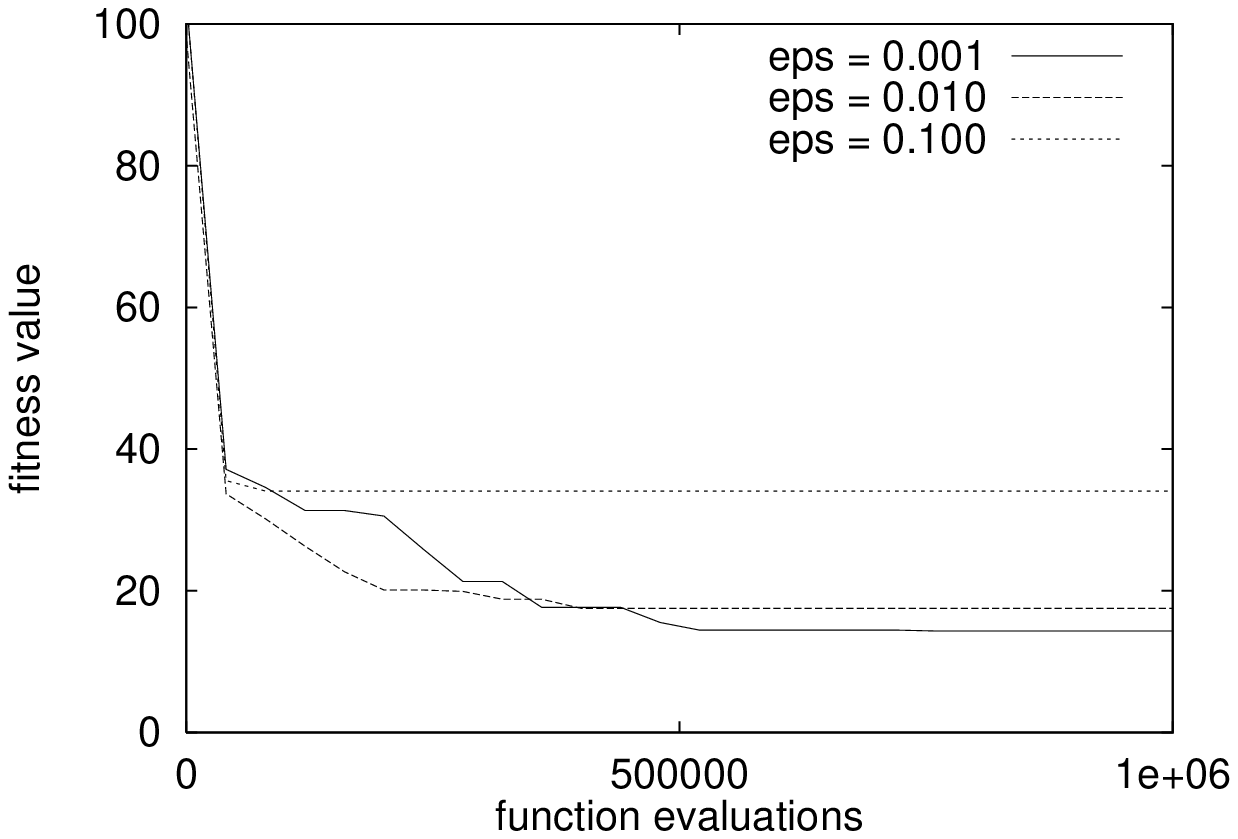}\\
Fig. 4. Comparing GP and SG-GP & &
Fig. 5. Influence of the learning rate\\
\end{tabular}
\end{center}
Fig. 4 shows the comparative behaviors of canonical GP and SG-GP
on the test identification problem.
The influence of the learning rate $\epsilon$ is depicted on Fig. 5, 
and of the mutation amplitude $\epsilon_m$ on 
Fig. 6.a. Overall, better results are obtained with a low learning rate
and a sufficiently large mutation amplitude; this can be interpreted
as a pressure toward the preservation of diversity.

\begin{center}
\begin{tabular}{ccc}
\includegraphics*[width=.4\linewidth]{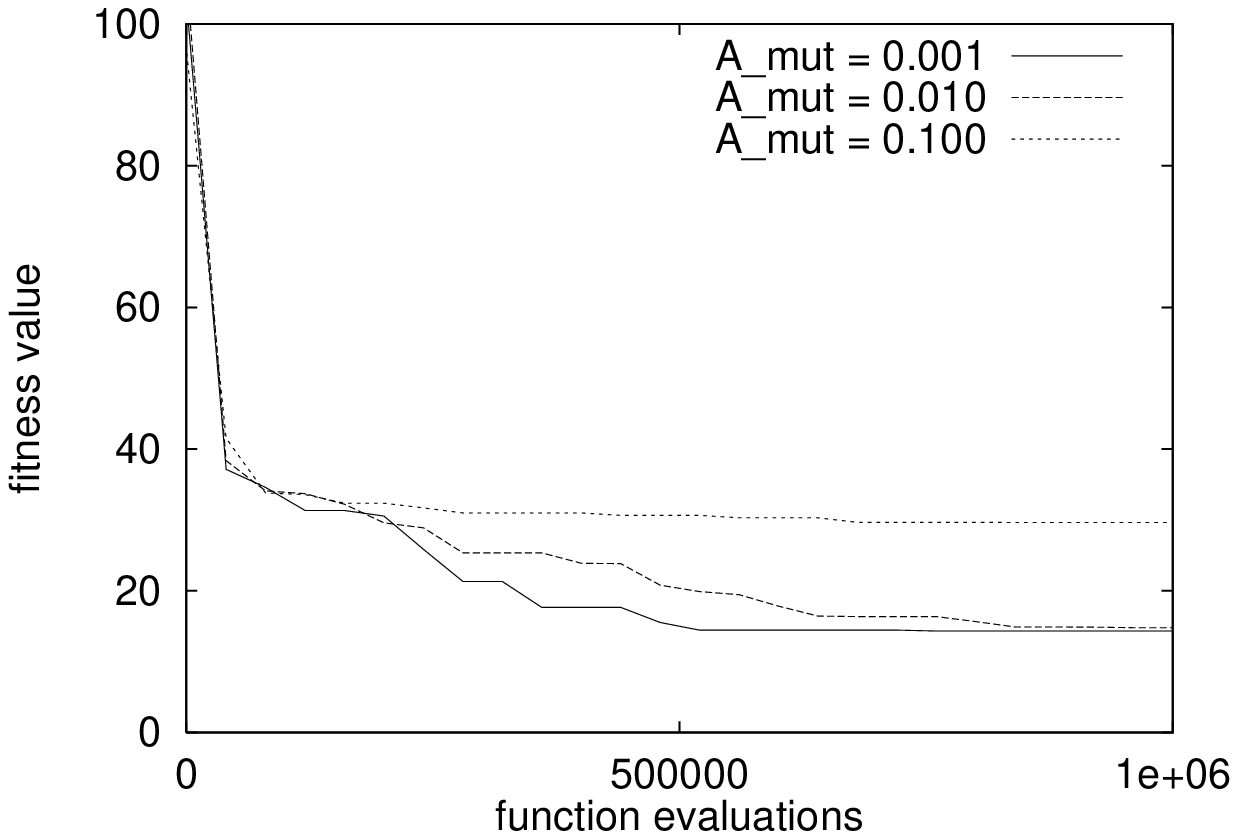} &\hspace{.2in} &
\includegraphics*[width=.4\linewidth]{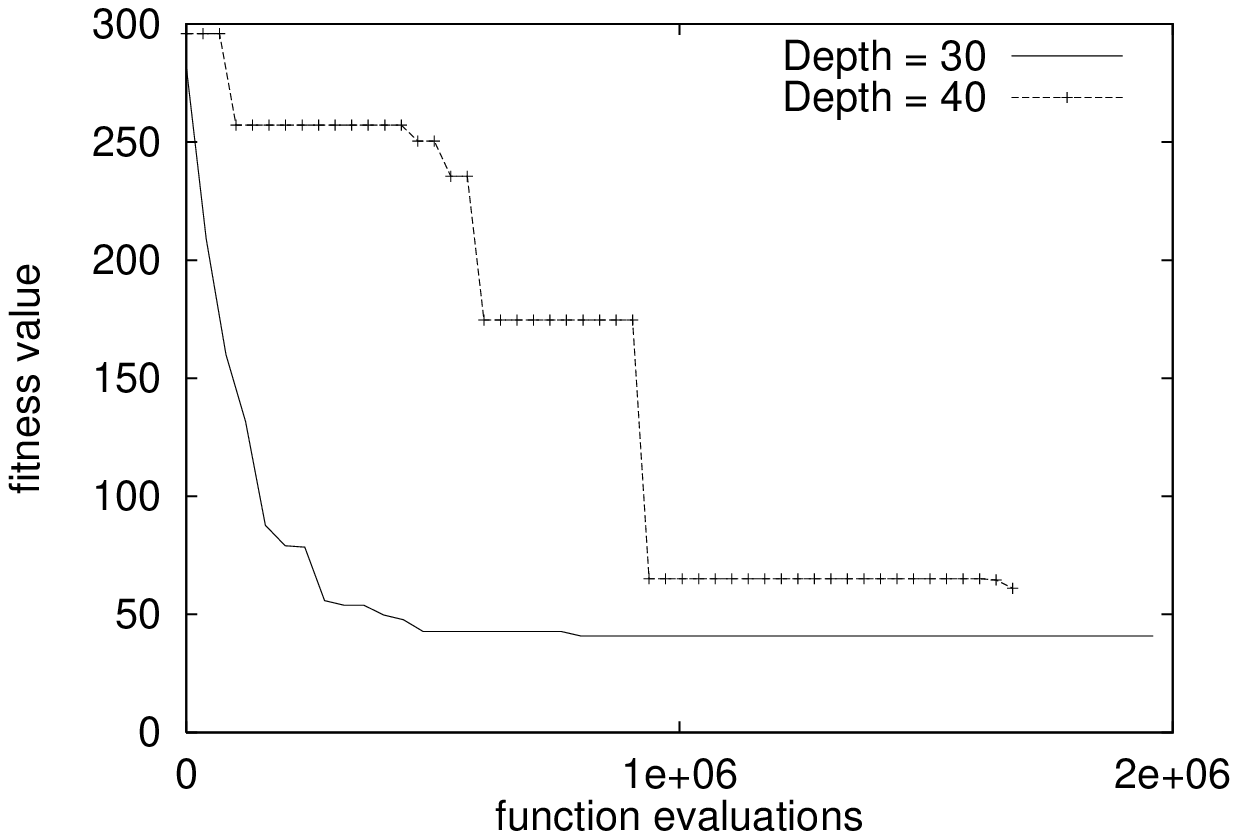}\\
(a) Mutation amplitude & & (b) Max. derivation depth\\
\end{tabular}\smallskip\\
Fig. 6. Parametric study of SG-GP
\end{center}

The maximum derivation depth allowed $D_{max}$  
is also a critical parameter. Too short, 
and the solution will be missed, too large, the search will take 
a prohibitively long time.
Fig.~6.b shows the solutions obtained with maximum derivation
depths of 30 and 40. As could have been expected, the solution is found 
faster for $D_{max} = 30$.

\begin{center}
\begin{tabular}{ccc}
\begin{tabular}{c}
\includegraphics*[width=.4\linewidth]{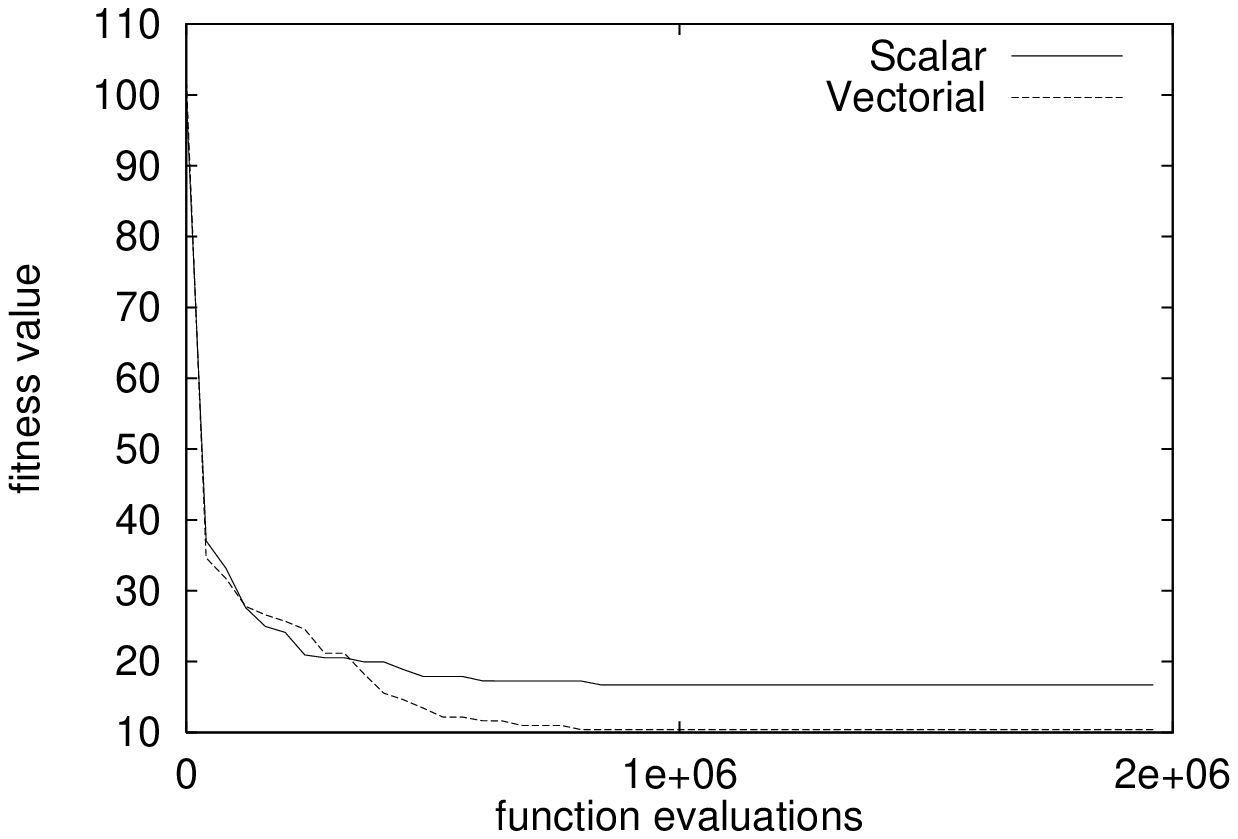}
\end{tabular} & \hspace*{.2in} &
{\small
\begin{tabular}{|l|cc|} \hline
& Scalar & Vectorial \\ \hline
Number of hits & 0 / 20 & 5 / 20\\
Av. best fitness & 16.7 & 10.3 \\
Std dev & $\pm$ 5 & $\pm$ 5\\\hline
\end{tabular}}\\
(a) Online performances & &
(b) Off line performances \\
\end{tabular}\smallskip\\
Fig. 7. Scalar vs Vectorial SG-GP
\end{center}

The advantage of using a vectorial distribution model against 
a scalar one, is illustrated on Fig. 7.a, as Vectorial SG-GP significantly 
improves on Scalar SG-GP. Table 7.b  points
out that vectorial SG-GP finds the target law (up to algebraic 
simplifications) after 2,000,000 fitness evaluations 
on 5 out of 20 runs, while no perfect match 
could be obtained with scalar SG-GP.

CFG-GP results, not shown here for space limitations, show that
even scalar SG-GP is more efficient than CFG-GP (see \cite{Ratle-Sebag:PPSN2000}
for more details).

\subsection{Resisting the bloat}
The most important experimental result is that 
SG-GP {\em does resist} the bloat, as it maintains 
an almost constant number of nodes. The average results over all
individuals and 20 runs is depicted on Fig. 9.a. 

In comparison is shown the number of nodes in GP 
(averaged on all individuals, but plotted for three 
typical runs for the sake of clarity). 
The individual size first drops in the 
first few generations; and after a while, it suddenly rises 
exponentially until the end of the run.
The drop is due to the fact that many trees created by 
crossover in the first generations are 
either trivial solutions (very simple trees) or infeasible 
solutions which are rejected. The rise occurs as soon as 
large feasible trees emerge in the population.

As noted by \cite{Banzhaf:book}, the size of the best individual 
is not correlated with the average size.
Figure 8.b shows the average size of the best-so-far individual. 
Interestingly, SG-GP maintains an almost constant size for the best individual,
which is slightly less than the average size. 
On the opposite, GP converges to a very small solution, despite the 
fact that most solutions are very large.

\begin{center}
\begin{tabular}{cc}
\includegraphics*[width=.4\linewidth]{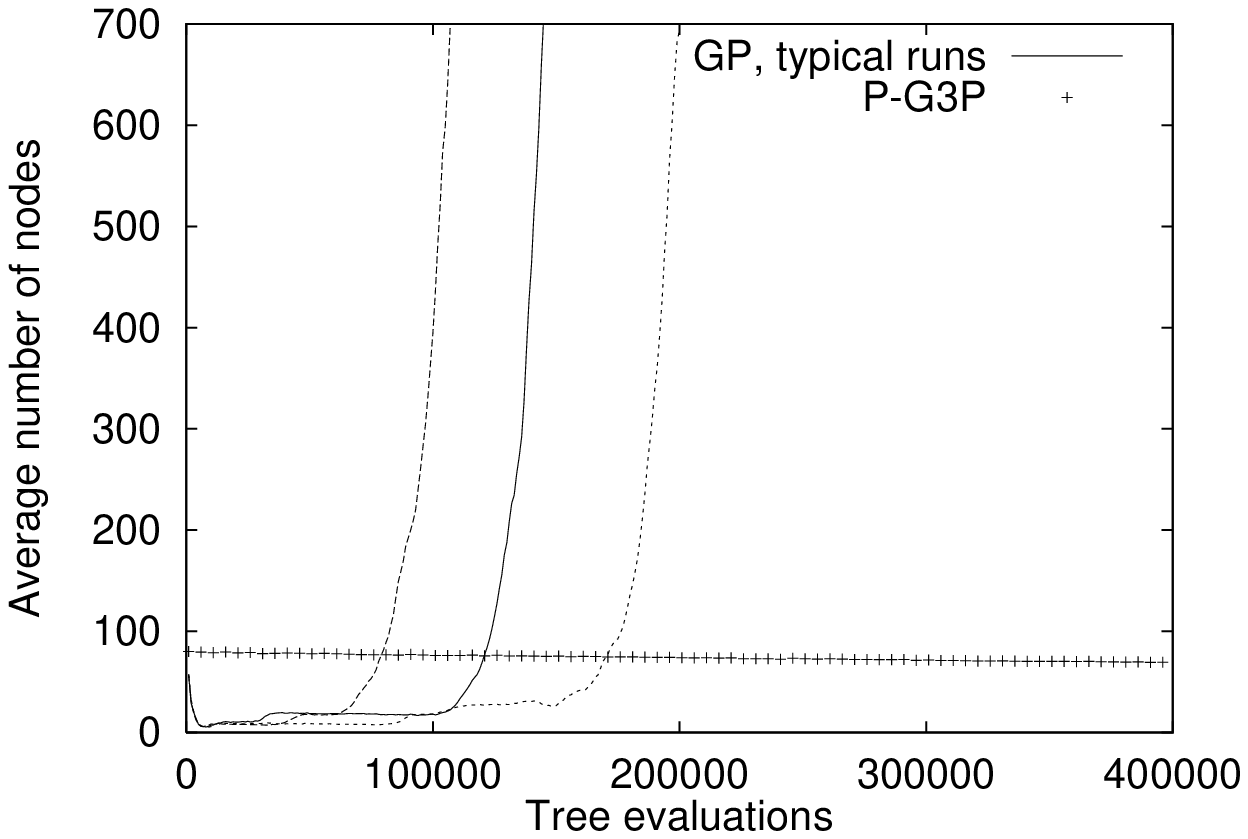} &
\includegraphics*[width=.4\linewidth]{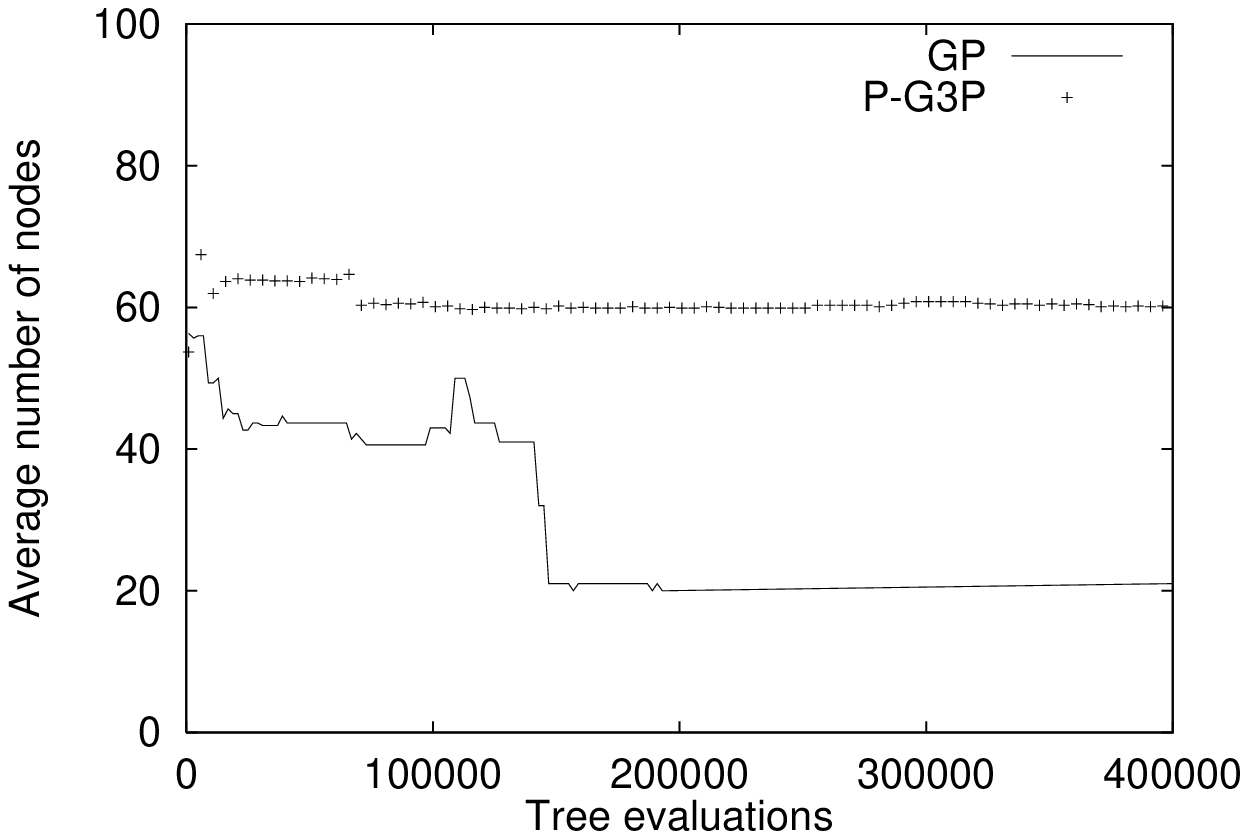}\\
(a) Average individual size &
(b) Best individual size\\
\end{tabular}\\
Fig. 8. Solution size with GP and SG-GP, averaged on 20 runs
\end{center}

\subsection{Identification and Generalization} 
As mentioned earlier on, SG-GP found the target law in 5 out of 
20 runs (up to algebraic simplifications). In most other runs, 
SG-GP converges toward a local optimum, a simplified expression of 
which is:
\begin{equation}
x(t)={\DIV{{2 * \MUL{t^2}{\DIV F K}}}{\ADD{\MUL{\DIV C K}{\DIV C K}}{2 * t^2}}}
\label{Solution135}
\end{equation}

This law is not on the path to the global optimum since the exponential  
is missing. However, the law closely fits  the 
training examples, at least from an engineering point of view (Fig. 9.a). 

\begin{center}
\begin{tabular}{cc}
\includegraphics*[width=.4\linewidth]{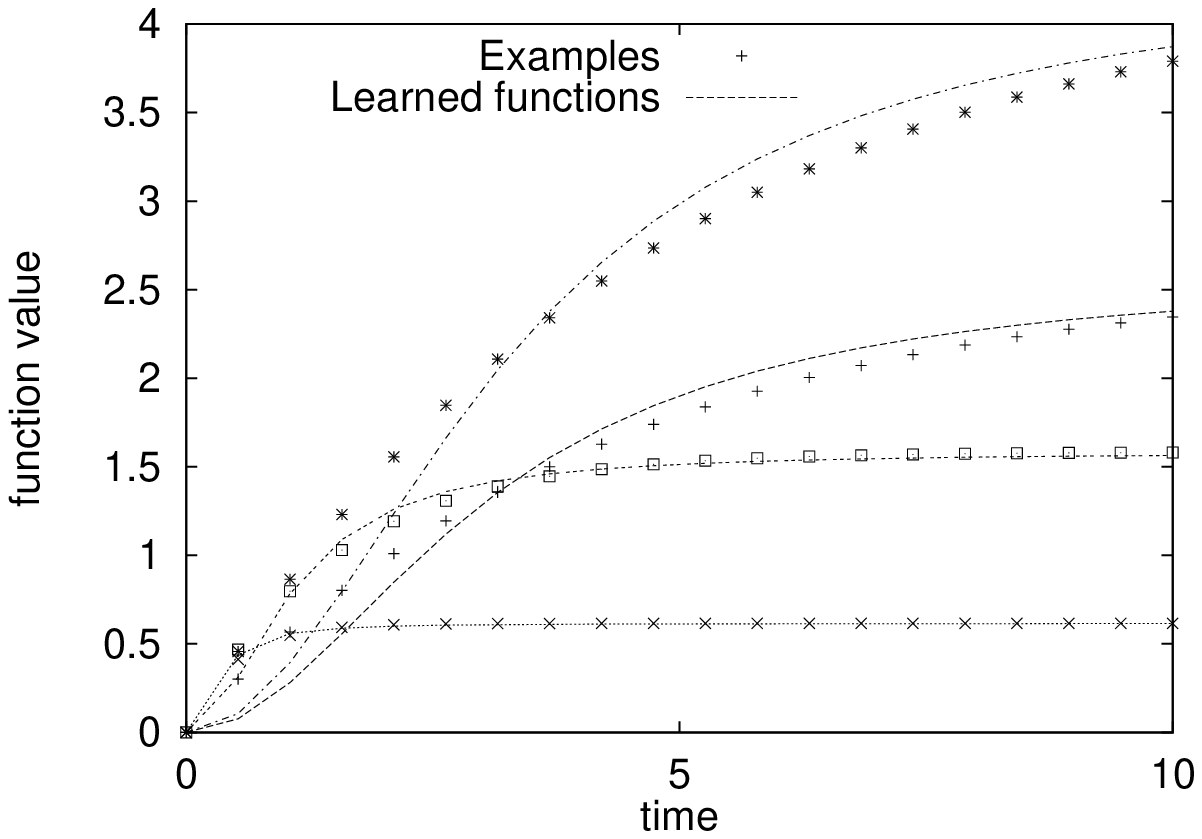} &
\includegraphics*[width=.4\linewidth]{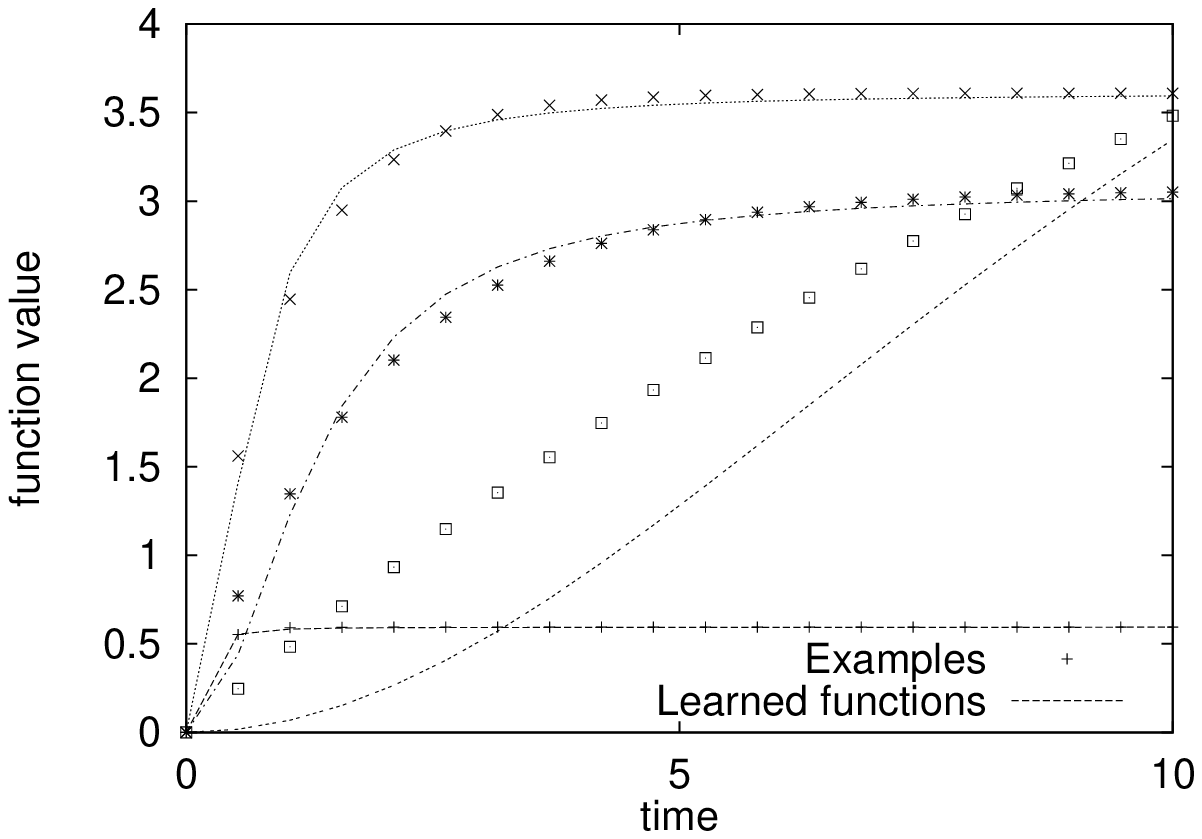}\\
(a) On training examples &
(b) On test examples\\
\end{tabular}\\
Fig. 9. Correct Identification and Generalization with SG-GP 
\end{center}

Even more important is the fact that SG-GP finds solutions which behaves
well on test examples, i.e. examples generated after the target law, which 
have not been considered during evolution.

By construction, the target law perfectly fits the test examples. 
But the non-optimal law (Eq \ref{Solution135}) 
also fits the test examples; the fit is quite
perfect is three out of four cases, and quite acceptable, 
from an engineer's point of view, in the last case.

\section{Conclusion}
In this paper was presented 
a novel
Genetic Programming scheme, combining grammar-based GP \cite{GruauPhD,Whigham95,Ratle-Sebag:PPSN2000} and  distribution-based evolution
\cite{Baluja:ICML95}, termed SG-GP for
Stochastic Grammar-based Genetic Programming. 
SG-GP differs from the PIPE system \cite{SalusSchmid98} as the 
distribution model used is based on stochastic grammars, which 
allows for overcoming one main limitation of GP, i.e.
the bloat phenomenon.

Intron growth was suggested to be unavoidable for  
program induction methods with fitness-based selection~\cite{LangdonPoli97a}. 

This conjecture
is infirmed by SG-GP results on a real-world like problem. Indeed, more 
intensive experiments are needed to see the limitations of the SG-GP scheme.

Still, SG-GP successfully resisted the intron growth on the problem
considered, in the following sense. \\
First of all, SG-GP shows good
identification abilities, as the target law was discovered
in 5 out of 20 runs, while it was never discovered by canonical GP.\\
Second, SG-GP shows good generalization abilities; even in the cases
where the target law was missed, the solutions found by SG-GP 
have good predictive accuracy on further examples
(not considered during 
the learning task).\\
Last, these identification and generalization tasks are 
successfully completed by exploring individuals with constant size.
No intron growth was observed; the overall memory requirements 
were lower by several orders of magnitude, than for canonical GP. 

These results suggest that intron growth is
not necessary to achieve efficient non parametric
learning in a fitness-based context, but might rather be a 
side effect of crossover-based evolution.

Further research is concerned with examining the actual
limitations of SG-GP through more intensive experimental validation. 
Efforts will be devoted to the parametric optimization problem
(find the constants) coupled with non-parametric optimization. 

%
%

\bibliographystyle{unsrt} 
\bibliography{LA_TOTALE}

\begin{thebibliography}{10}

\bibitem{Koza}
J.~R. Koza.
\newblock {\em Genetic Programming: On the Programming of Computers by means of
  Natural Evolution}.
\newblock MIT Press, Massachusetts, 1992.

\bibitem{Banzhaf:book}
W.~Banzhaf, P.~Nordin, R.E. Keller, and F.D. Francone.
\newblock {\em Genetic Programming --- An Introduction On the Automatic
  Evolution of Computer Programs and Its Applications}.
\newblock Morgan Kaufmann, 1998.

\bibitem{McKay95}
B.~McKay, M.J. Willis, and G.W. Barton.
\newblock Using a tree structures genetic algorithm to perform symbolic
  regression.
\newblock In {\em IEEE Conference publications, n. 414}, pages 487--492, 1995.

\bibitem{Duffy99}
J.~Duffy and J.~Engle-Warnick.
\newblock Using symbolic regression to infer strategies from experimental data.
\newblock In {\em Evolutionary Computation in Economics and Finance}. Springer
  Verlag, 1999.

\bibitem{Radcliffe}
N.~J. Radcliffe.
\newblock Equivalence class analysis of genetic algorithms.
\newblock {\em Complex Systems}, 5:183--20, 1991.

\bibitem{Janikow}
C.~Z. Janikow.
\newblock A knowledge-intensive genetic algorithm for supervised learning.
\newblock {\em Machine Learning}, 13:189--228, 1993.

\bibitem{Ratle-Sebag:PPSN2000}
A.~Ratle and M.~Sebag.
\newblock Genetic programming and domain knowledge: Beyond the limitations of
  grammar-guided machine discovery.
\newblock In M.~Schoenauer et~al., editor, {\em Proceedings of the $6^{th}$
  Conference on Parallel Problems Solving from Nature}, pages 211--220.
  Springer-Verlag, LNCS 1917, 2000.

\bibitem{GruauPhD}
F.~Gruau.
\newblock {\em Neural Network Synthesis using Cellular encoding and the Genetic
  Algorithm}.
\newblock PhD thesis, Ecole Normale Superieure de Lyon, 1994.

\bibitem{Whigham95}
P.A. Whigham.
\newblock Inductive bias and genetic programming.
\newblock In {\em IEEE Conference publications, n. 414}, pages 461--466, 1995.

\bibitem{LangdonPoli97a}
W.~B. Langdon and R.~Poli.
\newblock Fitness causes bloat.
\newblock In {\em Soft Computing in Engineering Design and Manufacturing},
  pages 13--22. Springer Verlag, 1997.

\bibitem{Baluja:ICML95}
S.~Baluja and R.~Caruana.
\newblock Removing the genetics from the standard genetic algorithms.
\newblock In A.~Prieditis and S.~Russel, editors, {\em Proceedings of the
  12$^{th}$ International Conference on Machine Learning}, pages 38--46. Morgan
  Kaufmann, 1995.

\bibitem{SalusSchmid98}
R.~Salustowicz and J.~Schmidhuber.
\newblock Evolving structured programs with hierarchical instructions and skip
  nodes.
\newblock In J.~Shavlik, editor, {\em Proceedings of the 15$^{th}$
  International Conference on Machine Learning}, pages 488--496. Morgan
  Kaufmann, 1998.

\bibitem{Montana95}
David~J. Montana.
\newblock Strongly typed genetic programming.
\newblock {\em Evolutionary Computation}, 3(2):199--230, 1995.

\bibitem{Ryan98}
C.~Ryan, J.J. Collins, and M.~O'Neill.
\newblock Grammatical evolution: Evolving programs for an arbitrary language.
\newblock In W.~Banzhaf, R.~Poli, M.~Schoenauer, and T.C. Fogarty, editors,
  {\em Genetic Programming, First European Workshop, EuroGP98}, volume LNCS
  1391, pages 83--96. Springer Verlag, 1998.

\bibitem{Sebag:PPSN98}
M.~Sebag and A.~Ducoulombier.
\newblock Extending population-based incremental learning to continuous search
  spaces.
\newblock In Th. B{\"a}ck, G.~Eiben, M.~Schoenauer, and H.-P. Schwefel,
  editors, {\em Proceedings of the $5^{th}$ Conference on Parallel Problems
  Solving from Nature}, pages 418--427. Springer Verlag, 1998.

\bibitem{Larranaga}
P.~Larranaga and J.~A. Lozano.
\newblock {\em Estimation of Distribution Algorithms. A New Tool for
  Evolutionary Computation}.
\newblock Kluwer Academic Publishers, 2001.

\bibitem{NBF97}
P.~Nordin, W.~Banzhaf, and F.D. Francone.
\newblock Introns in nature and in simulated structure evolution.
\newblock In D.~Lundh, B.~Olsson, and A.~Narayanan, editors, {\em Biocomputing
  and Emergent Computation}, pages 22--35. World Scientific, 1997.

\bibitem{parsimony}
Byoung-Tak Zhang and Heinz M{\"u}hlenbein.
\newblock Balancing accuracy and parsimony in genetic programming.
\newblock {\em Evolutionary Computation}, 3(1):17--38, 1995.

\bibitem{Ward85}
I.M. Ward.
\newblock {\em Mechanical Properties of Solid Polymers}.
\newblock Wiley, Chichester, 1985.

\end{thebibliography}
\end{document}